\documentclass[10pt,twocolumn,letterpaper]{article}
\usepackage[pagenumbers]{cvpr}
\usepackage{amsmath,amssymb,amsthm}
\usepackage{graphicx,booktabs,array,multirow}
\usepackage{enumitem,microtype}
\usepackage{xcolor}
\usepackage{threeparttable}

\usepackage{tikz}
\usetikzlibrary{arrows.meta,calc,positioning,patterns,patterns.meta,decorations.pathreplacing}

\definecolor{cInk}{HTML}{1B1F27}     % lettering, badges
\definecolor{cText}{HTML}{434B59}    % secondary lettering
\definecolor{cMuted}{HTML}{6B7482}   % muted lettering
\definecolor{cGreyM}{HTML}{C9CED6}   % rules, outlines
\definecolor{cGreyL}{HTML}{ECEEF2}   % Table 1: heads of column groups, tags, the track of a data bar (the grey of the tracks in the figures)
\definecolor{cBT}{HTML}{1F6FE5}      % Event BT
\definecolor{cFC}{HTML}{F08A00}      % Full Channel
\definecolor{cTruth}{HTML}{0FA58F}   % the truth; readouts with additional truth information (*)
\definecolor{cFrozen}{HTML}{B5379F}  % the starting model (Frozen)
\definecolor{cSelfLine}{HTML}{4E5666} % the self control

\newcommand{\introVS}{0pt}         % Fig. 1: space added under its caption
\newcommand{\methodVS}{0pt}        % Fig. 2
\newcommand{\taskVS}{0pt}         % Fig. 3
\newcommand{\tableVS}{0pt}         % Table 1
\newcommand{\pushVS}{0pt}          % Fig. 4
\newcommand{\robustVS}{0pt}        % Fig. 6
\newcommand{\eventsVS}{0pt}        % Fig. 5

\makeatletter
\newcommand{\RefsTailA}[3]{% #1: the word up to its last \penalty0\relax (or empty), #2: the rest of it, #3: the year
  \setbox\z@\hbox{#2, #3\ 0}%
  \ifdim\wd\z@<60pt
    \setbox\tw@\hbox{#1#2, #3\ 0}%
    \ifdim\wd\tw@<60pt \nobreak\fi\ \mbox{#1#2, #3}%
  \else
    \ #1\mbox{#2, #3}%
  \fi}
\newcommand{\RefsTailB}[1]{% #1: the last word of a group that is followed by ", YYYY."
  \setbox\z@\hbox{#1, 0000.\ 0}%
  \ifdim\wd\z@<60pt \nobreak\fi\ \mbox{#1}}
\makeatother
\ExplSyntaxOn
\tl_new:N \l__refs_entry_tl
\cs_new_protected:Npn \RefsTieTail #1
  {
    \tl_set:Nn \l__refs_entry_tl {#1}
    \regex_replace_all:nnN { \c{penalty} (\d+) \  } { \c{penalty} \1 \c{relax} } \l__refs_entry_tl
    \regex_replace_once:nnNF
      { \  ( (?:\c[^BES].)*? ) ( (?:\c[^BESC].)+ ) \, \  (\d\d\d\d\.) (\ *) \Z }
      { \c{RefsTailA}\cB\{\1\cE\}\cB\{\2\cE\}\cB\{\3\cE\}\4 } \l__refs_entry_tl
      {
        \regex_replace_once:nnNF
          { \  ( (?:\c[^BES].)+ ) (\cE.) \, \  (\d\d\d\d\.) (\ *) \Z }
          { \c{RefsTailB}\cB\{\1\cE\}\2\,\c{nobreak}\ \3\4 } \l__refs_entry_tl
          { \regex_replace_once:nnN { \, \  (\d\d\d\d\.) (\ *) \Z } { \,\c{nobreak}\ \1\2 } \l__refs_entry_tl }
      }
    \tl_use:N \l__refs_entry_tl
  }
\ExplSyntaxOff
\makeatletter
\AtBeginDocument{%
  \@ifpackageloaded{backref}{%
    \def\BR@@lbibitem[#1]#2#3\par{\let\backrefprint\BR@backrefprint
      \BRorg@bibitem[{#1}]{#2}\RefsTieTail{#3}\BR@backref{#2}}}{}}
\makeatother
\newcommand{\E}{\mathbb{E}}
\newcommand{\ind}{\mathbf{1}}
\newcommand{\bt}{\mathrm{BT}}

\newcommand{\na}{\textemdash}
\colorlet{tabgrey}{cMuted}             % differences whose interval includes zero; secondary lettering (the muted lettering of the figures)
\newlength{\tabrowht}% the height of a table row (\arraystretch times \baselineskip), set in the table
\newcommand{\rmark}[1]{\makebox[6.4pt][c]{\tikz[baseline=-0.62ex]{\csname rmark@#1\endcsname}}\hspace{3.2pt}}
\makeatletter
\@namedef{rmark@bt}{\fill[cBT] (0,0) circle (2.35pt);}
\@namedef{rmark@fc}{\fill[cFC] (-2.05pt,-2.05pt) rectangle (2.05pt,2.05pt);}
\@namedef{rmark@mle}{\fill[cTruth] (-2.6pt,-2.1pt) -- (2.6pt,-2.1pt) -- (0,2.5pt) -- cycle;}
\@namedef{rmark@self}{\fill[cSelfLine] (0,-2.7pt) -- (2.4pt,0) -- (0,2.7pt) -- (-2.4pt,0) -- cycle;}
\@namedef{rmark@frozen}{\draw[cFrozen, line width=1.2pt, dash pattern=on 0.9pt off 1.5pt] (-3.1pt,0) -- (3.1pt,0);}
\newcommand{\ricon}[1]{\hspace{2.4pt}\raisebox{-0.9pt}{#1}}
\newcommand{\tbadge}[1]{\raisebox{-1.6pt}{#1}}
\newcommand{\tblock}[2]{{\fontsize{9.5}{10}\selectfont\bfseries #1}\hspace{4pt}\textbf{#2}}
\newlength{\tab@tmp}\newlength{\tab@off}\newlength{\tab@unit}\newcount\tab@n\newcount\tab@sum
\newlength{\tabpadx}
\newlength{\tabov}
\newcommand{\tabmeasure}[2]{%
  \@ifundefined{tabw@#1}{\expandafter\newlength\csname tabw@#1\endcsname}{}%
  \setlength{\csname tabw@#1\endcsname}{0pt}%
  \@for\tab@v:=#2\do{\settowidth{\tab@tmp}{\tab@v}%
    \ifdim\tab@tmp>\csname tabw@#1\endcsname\setlength{\csname tabw@#1\endcsname}{\tab@tmp}\fi}}
\newcommand{\tabgrid}[2]{%
  \setlength{\tab@tmp}{\dimexpr\textwidth-2\tabpadx\relax}\tab@n=0 \tab@sum=0
  \@for\tab@v:=#2\do{\advance\tab@n by 1 \advance\tab@sum by \tab@v\relax
    \addtolength{\tab@tmp}{-\csname tabw@#1\the\tab@n\endcsname}}%
  \advance\tab@n by 1 \addtolength{\tab@tmp}{-\csname tabw@#1\the\tab@n\endcsname}%
  \setlength{\tab@unit}{\dimexpr\tab@tmp/\tab@sum\relax}%
  \ifdim\tab@unit<0.2pt\PackageWarning{tables}{Table grid #1 too narrow}\fi
  \tab@n=0
  \@for\tab@v:=#2\do{\advance\tab@n by 1
    \@ifundefined{tabg@#1\the\tab@n}{\expandafter\newlength\csname tabg@#1\the\tab@n\endcsname}{}%
    \setlength{\csname tabg@#1\the\tab@n\endcsname}{\tab@v\tab@unit}}%
  \typeout{Table grid #1: unit \the\tab@unit}}
\newcommand{\tabw}[1]{\csname tabw@#1\endcsname}
\newcommand{\tabg}[1]{\csname tabg@#1\endcsname}
\newcommand{\tabbanner}[6]{%
  \setlength{\tab@tmp}{2\tabov}\tab@n=#2\relax
  \loop\addtolength{\tab@tmp}{\csname tabw@#1\the\tab@n\endcsname}%
    \ifnum\tab@n<#3\relax\addtolength{\tab@tmp}{\csname tabg@#1\the\tab@n\endcsname}\advance\tab@n by 1 \repeat
  \hspace*{-\tabov}\rlap{\tikz[baseline=(g.base)]{\node[inner xsep=0pt, inner ysep=1.7pt,
    text height=1.75ex, text depth=0.45ex] (g) {\makebox[\tab@tmp][l]{\hspace{\tabov}\ifx\relax#4\relax\else\tbadge{#4}\hspace{2.6pt}\fi
    \textbf{#5}\ifx\relax#6\relax\else\hspace{3.4pt}{\color{cText}#6}\fi}};
    \draw[cInk, line width=0.4pt] ([xshift=\tabov]g.south west) -- ([xshift=-\tabov]g.south east);}}}
\newcommand{\tabbannersplit}[8]{%
  \setlength{\tab@tmp}{2\tabov}\setlength{\tab@off}{\tabov}\tab@n=#2\relax
  \loop\addtolength{\tab@tmp}{\csname tabw@#1\the\tab@n\endcsname}%
    \ifnum\tab@n<#3\relax\addtolength{\tab@off}{\csname tabw@#1\the\tab@n\endcsname}%
      \addtolength{\tab@off}{\csname tabg@#1\the\tab@n\endcsname}\fi
    \ifnum\tab@n<#4\relax\addtolength{\tab@tmp}{\csname tabg@#1\the\tab@n\endcsname}\advance\tab@n by 1 \repeat
  \tab@n=#3\advance\tab@n by -1
  \addtolength{\tab@off}{-0.5\csname tabg@#1\the\tab@n\endcsname}%
  \hspace*{-\tabov}\rlap{\tikz[baseline=(g.base)]{\node[inner xsep=0pt, inner ysep=1.7pt,
    text height=1.75ex, text depth=0.45ex] (g) {\makebox[\tab@tmp][l]{\hspace{\tabov}\tbadge{#5}\hspace{2.6pt}\textbf{#6}\hspace{3.4pt}{\color{cText}#7}}};
    \draw[cInk, line width=0.4pt] ([xshift=\tabov]g.south west) -- ([xshift=\tab@off-2pt]g.south west);
    \draw[cInk, line width=0.4pt] ([xshift=\tab@off+2pt]g.south west) -- ([xshift=-\tabov]g.south east);
    \node[anchor=base west, inner sep=0pt] at ([xshift=\tab@off+2pt]g.base west) {\color{cText}#8};}}}
\newcommand{\tabgroup}[1]{\ifx\relax#1\relax\else\textit{#1}\fi}
\newcommand{\tabsig}[1]{\textbf{\boldmath #1}}            % Event BT worse than every other readout (block a): bold
\newcommand{\tabns}[1]{\textcolor{tabgrey}{#1}}           % a difference whose interval includes zero
\newlength{\tabpad}\newlength{\tabgut}
\newcommand{\tabsetcol}[3]{%
  \@ifundefined{tabvw@#1}{\expandafter\newlength\csname tabvw@#1\endcsname
                          \expandafter\newlength\csname tabtw@#1\endcsname}{}%
  \setlength{\csname tabvw@#1\endcsname}{0pt}%
  \@for\tab@v:=#2\do{\settowidth{\tab@tmp}{\tab@v}%
    \ifdim\tab@tmp>\csname tabvw@#1\endcsname\setlength{\csname tabvw@#1\endcsname}{\tab@tmp}\fi}%
  \setlength{\csname tabtw@#1\endcsname}{\dimexpr 2\csname tabvw@#1\endcsname+4\tabpad+\tabgut\relax}%
  \@for\tab@v:=#3\do{\settowidth{\tab@tmp}{\tab@v}%
    \ifdim\dimexpr\tab@tmp+2\tabpad\relax>\csname tabtw@#1\endcsname\setlength{\csname tabtw@#1\endcsname}{\dimexpr\tab@tmp+2\tabpad\relax}\fi}}
\newcommand{\tab@val}[3]{\makebox[#1][c]{\ifnum#2=1 #3\else\textcolor{tabgrey}{#3}\fi}}
\newcommand{\tabpair}[5]{\makebox[\csname tabtw@#1\endcsname][c]{%
  \tab@val{\dimexpr\csname tabvw@#1\endcsname+2\tabpad\relax}{#2}{#3}\hspace{\tabgut}%
  \tab@val{\dimexpr\csname tabvw@#1\endcsname+2\tabpad\relax}{#4}{#5}}}
\newcommand{\tabwide}[3]{\tab@val{\csname tabtw@#1\endcsname}{#2}{#3}}
\newcommand{\tabpairw}[1]{\csname tabtw@#1\endcsname}  % the width of a column of pairs
\makeatother

\definecolor{cvprblue}{rgb}{0.21,0.49,0.74}
\usepackage[pagebackref,breaklinks,colorlinks,allcolors=cvprblue]{hyperref}
\hypersetup{pdftitle={Bradley\textendash Terry Training Can Distort Outcome Probabilities of Initially Accurate Language World Models},pdfauthor={Xinze Shi, Litian Zhang, Zhenjie Su, Zelong Yu, Yuting Zhang, Chaozhuo Li, Qihua Feng, Zifeng Kang, Zhengxuan Hu}}
\title{Bradley--Terry Training Can Distort Outcome Probabilities\\of Initially Accurate Language World Models}
\author{Xinze Shi\textsuperscript{1}, \quad Litian Zhang\textsuperscript{1}, \quad Zhenjie Su\textsuperscript{1}, \quad Zelong Yu\textsuperscript{2}, \quad Yuting Zhang\textsuperscript{1}\\[2pt]
Chaozhuo Li\textsuperscript{3}, \quad Qihua Feng\textsuperscript{4}, \quad Zifeng Kang\textsuperscript{1}, \quad Zhengxuan Hu\textsuperscript{5}\\[4pt]
{\small \textsuperscript{1}Beijing University of Posts and Telecommunications\qquad \textsuperscript{2}Beihang University}\\
{\small \textsuperscript{3}Microsoft\qquad \textsuperscript{4}Beijing Institute of Technology\qquad \textsuperscript{5}Queen Mary, University of London}}

\begin{document}
\maketitle
\begin{abstract}
Comparative feedback, in which a judge says which of two candidate outcomes is closer to what happened, is a practical way to train language world models that assign probabilities to an action's outcomes. We show that, fitted with a Bradley--Terry objective, it need not teach the true outcome probabilities and can distort accurate ones. On a simulated robot push with enumerated outcome distributions, this objective takes a model that is accurate on unseen layouts (hazard error 2.4 percentage points, pp) and moves 33 to 43~pp of its probability onto outcomes that none of a condition's enumerated branches produces. The cause is the objective, not the comparisons: its optimum, computed before training, predicts where the model ends up, and a likelihood that knows the judge's rule learns from the same comparisons without the shift. The shift is robust: it recurs on unseen conditions, under five candidate laws and a threefold denser enumeration, and about 29~pp remain without the physically impossible events. Its cost is conditional: with those events it inflates hazard probabilities, and a threshold inherited from the pre-feedback model costs about 3~pp of utility per decision on trained and 1 to 2~pp on unseen conditions, through false rejects; without them the utility loss averages below 1~pp over six seeds, though a small regret in action choice remains. Audits of readout, response and decision trace the effect from probabilities to actions.
\end{abstract}

\section{Introduction}
\begin{figure}[t]
\centering\includegraphics[width=\columnwidth]{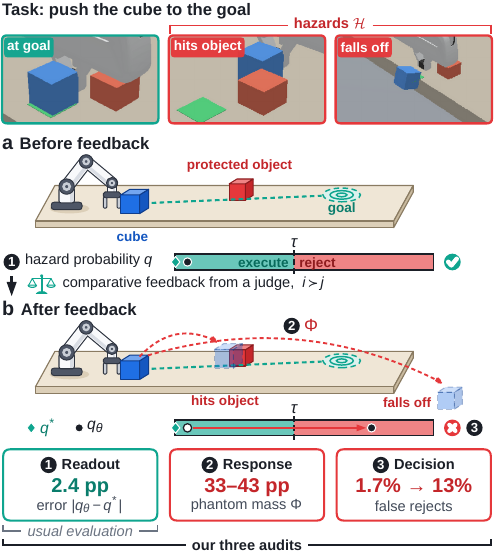}
\caption{Comparative feedback on a robot push with two hazards~$\mathcal H$. (a)~Before feedback, the hazard probability~$q$ of a language world model is close to the truth and below a fixed threshold~$\tau$: the push is executed. (b)~After feedback, 33 to 43~pp of probability ($\Phi$) sit on outcomes that never occur in that condition, and safe pushes are held back. \emph{Bottom}: the usual evaluation against our three audits. Scenes and strips schematic, numbers measured; frames from different rollouts.}
\label{fig:intro}
\vspace{\introVS}
\end{figure}

World models support planning by predicting the consequences of candidate actions~\cite{worldmodels,dreamer}. A language world model does so in text: it describes a finite set of possible outcomes and assigns each a probability. When some outcomes are hazardous, their total probability can be compared with a fixed threshold to decide whether to execute or reject an action; \cref{fig:intro}(a) shows this workflow for a robot push.

Such a model needs supervision, and the frequencies of rare outcomes are hard to observe directly. Comparative feedback, familiar from preference learning~\cite{humanpref,dpo}, is a practical substitute: an annotator or a simulator compares two candidate outcome descriptions under a criterion such as realism. It is natural to expect that a model trained to agree with truthful comparisons also learns how often each outcome occurs. We show that it need not. A preference between two descriptions does not reveal how frequently either outcome occurs, and an objective fitted to preferences has an optimum of its own. Feedback can therefore move an already accurate model away from the true probabilities, and the decisions that depend on them move with it.

We establish this on a simulated robot push in which enumerating hidden friction, control latency and position error gives the outcome distribution of every condition (\cref{fig:intro}). A language model trained on such distributions is accurate on unseen layouts (hazard error 2.4 percentage points, pp). Continuing its training with Bradley--Terry feedback from a truthful judge moves 33 to 43~pp of its probability onto outcomes that none of a condition's enumerated branches produces. More comparisons do not remove the shift, because it is where the objective's optimum lies: the optimum, computed before training, predicts the endpoint, and a likelihood of the same comparisons that knows the judge's rule shows no such shift. The shift recurs on unseen conditions, under five candidate laws, against a denser enumeration and without the four physically impossible events. Its cost is conditional. With those events it inflates the hazard probability, and a threshold inherited from the pre-feedback model earns 3.2~pp less utility per decision than the model trained with the judge-aware likelihood, and 1.3 to 1.8~pp less on unseen conditions; re-tuning the threshold recovers the loss, but needs the ground truth that comparative feedback was meant to replace. Without those events the utility loss averages below 1~pp, and a small regret in action choice remains.

\begin{figure*}[t]
\centering
\includegraphics[width=\textwidth]{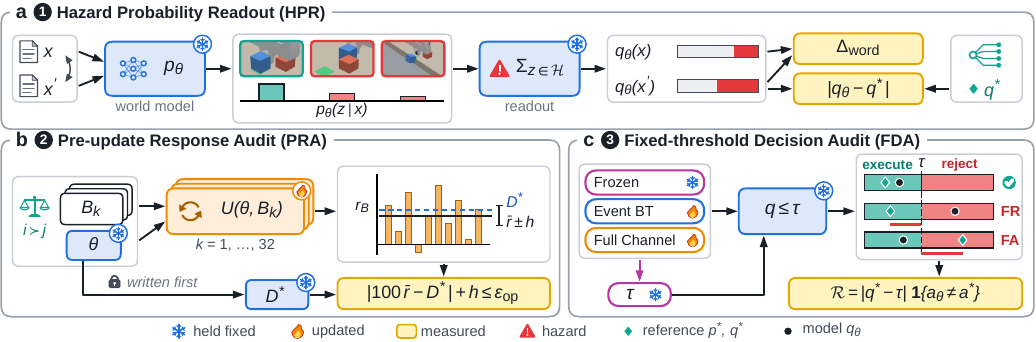}
\caption{The three audits: what each takes as input, holds fixed (snowflake) or updates (flame), and measures (yellow; \cref{eq:word,eq:operational,eq:regret}). $q$ is a probability and \mbox{1~pp $=0.01$}; $D^*$, $h$ and $\epsilon_{\mathrm{op}}$ are in pp. Sketches are schematic: the red part of a bar in (a) is the hazard probability, each small bar in (b) is the response to one batch, and each strip in (c) is one condition under the threshold $\tau$ chosen for Frozen, with a red line for its regret; the pictures in (a) are simulation frames.}
\label{fig:method}
\vspace{\methodVS}
\end{figure*}

We trace the effect with three audits, each with fixed information access (\cref{fig:method}). \emph{Hazard Probability Readout} (HPR) measures a frozen model's hazard probability against the reference and under rewording. \emph{Pre-update Response Audit} (PRA) scores forecasts of what feedback will do, committed before the feedback is observed, for a single update and for the converged endpoint. \emph{Fixed-threshold Decision Audit} (FDA) evaluates the resulting actions under a common threshold, inherited or re-tuned. The calculations are standard~\cite{influence,tracin,bradley,hunter}; the audits fix when, and with what information, they are evaluated, which separates an effect of the objective from drift, noise and hindsight.

Mean probability error on fixed descriptions, the usual evaluation, can show that probabilities are off, but not why or at what cost. This is an existence result in one controlled setting, where only the objective varies: the truth is enumerated exactly, the judge is truthful with a stated rule, the starting model already fits the truth, and Full Channel, a likelihood that knows that rule, learns from the same comparisons; the setting has one text-described task, one model and one judge. The optimum behind it, however, depends only on the outcome distribution, the candidate law and the judge (\cref{eq:bstar}), not on the model or its input.
\par
Our contributions are:\nopagebreak
\begin{itemize}[leftmargin=*,nosep]
\item An existence result: Bradley--Terry feedback moves 33 to 43~pp of an initially accurate language world model's probability outside the reference support, while a judge-aware likelihood, outcome supervision and a self-generated control stay at or below 1.4~pp (\cref{sec:exp-main}).
\item An explanation checked in advance: the objective's optimum, computed before training, forecasts where the model ends up and which outcomes gain probability, whereas a single update cannot be forecast much better than by predicting no change (\cref{sec:exp-response}).
\item A conditional decision cost: under an inherited threshold the shift costs 3.2~pp of utility per decision on trained and 1.3 to 1.8~pp on unseen conditions, through false rejects, a loss that re-tuning removes and a higher hazard cost can reverse~(\cref{sec:exp-decision}).
\item Robustness: the shift survives on the feasible events over six seeds, on a denser reference and under five candidate laws; without the impossible events the mean utility loss falls below 1~pp while a small action regret \mbox{remains}~(\cref{sec:exp-robust}).
\item Three audits with fixed information access that measure these effects (\cref{sec:method}), illustrated on synthetic structures.
\end{itemize}

\section{Related work}
% Figure 2 (the three audits of Sec. 3) spans both columns. It is declared here, on page 2, so that it is set at
% the top of page 3, where Sec. 3 begins.
\textbf{Preference learning.}
Pairwise judgments extend supervision beyond demonstrations~\cite{humanpref,instruct}; DPO and related objectives optimize a policy directly from preference pairs~\cite{dpo,ipo,simpo,kto,tpo}, many through the Bradley--Terry model~\cite{bradley,hunter}, and a proxy can improve while its target does not~\cite{overopt}. Closest to us, Siththaranjan et al.~\cite{hiddencontext} prove that when preferences depend on a context the learner does not observe, the Bradley--Terry optimum for uniformly drawn pairs ranks the alternatives by their Borda count, their average probability of winning, not by expected utility. Event BT is a fit of this kind, with the realized outcome as the context, so its scores reflect how often an outcome wins a comparison, not how often it occurs. That result fixes only their order. We add (i)~their \emph{values}, which a probability readout uses: $b^*$ (\cref{eq:bstar}) places 37.9~pp on average on outcomes a condition never produces; (ii)~\emph{non-uniform candidates}, drawn here mostly from the model's own prediction, under which even the order departs from the Borda count in 107 of 153 conditions (Supplementary \S A.3); we test five laws, the uniform one included; and (iii)~\emph{training and decisions}: they fit reward models; we train the world model, forecast its endpoint before training and measure its cost to a threshold decision.

\textbf{World models and planning.}
Learned dynamics support imagination, planning and model-generated experience~\cite{worldmodels,dreamer,dreamerv3,muzero,pets,mbpo}, and offline methods counter model error with pessimism or constraints~\cite{mopo,morel,cpo}. Language models connect generated descriptions to planning~\cite{rap,react,saycan}, and NWM predicts visual futures~\cite{nwm}. We audit finite-event language models grounded in a physics simulator~\cite{maniskill3,sapien}; visual inputs need their own evidence.

\textbf{Predictive uncertainty.}
Calibration and dataset-shift research examine whether confidence matches correctness and stays reliable~\cite{calibration,shift,ensembles}. For language models, self-evaluation, verbalized confidence and semantic uncertainty study how uncertainty is elicited~\cite{know,askcal,semantic}, and risk-controlling prediction sets and conformal risk control give finite-sample guarantees under their assumptions~\cite{rcps,crc}. Our response audit scores a committed forecast against the updates that follow; a bound on it does not bound error relative to environmental probabilities.

\section{Problem and audit method}
\label{sec:method}
\subsection{Comparative feedback and its optimum}
\label{sec:formulation}
Let $x$ describe a state and a candidate action, and let $\mathcal Z=\{1,\ldots,K\}$ be a finite set of mutually exclusive, exhaustive outcomes. A world model with parameters $\theta$ predicts $p_\theta(z\mid x)$; the environment follows the reference distribution $p^*(z\mid x)$. For a fixed hazard subset $\mathcal H\subseteq\mathcal Z$, the hazard probabilities are
\begin{equation}
q_\theta(x)=\sum_{z\in\mathcal H}p_\theta(z\mid x),\qquad
q^*(x)=\sum_{z\in\mathcal H}p^*(z\mid x).
\label{eq:hazard}
\end{equation}

\textbf{Comparative feedback.}
A comparison presents two candidate outcomes $i,j\in\mathcal Z$ for the same $x$, drawn from a candidate law $r(\cdot\mid x)$, and a judge returns a label $u$: $i$ wins, $j$ wins or a tie. We consider a truth judge. An outcome $y\sim p^*(\cdot\mid x)$ is realized, and the candidate nearer to $y$ under a fixed distance between outcomes wins. Two objectives turn the label into a loss. \emph{Event BT} applies the Bradley--Terry model~\cite{bradley} to the model's own event probabilities,
\begin{equation}
\ell_{\bt}=-\log\frac{p_\theta(w\mid x)}{p_\theta(w\mid x)+p_\theta(l\mid x)},
\label{eq:bt}
\end{equation}
for winner $w$ and loser $l$; a tie contributes nothing. \emph{Full Channel} is the likelihood of the label itself, with the model's distribution standing in for that of the outcome:
\begin{equation}
\ell_{\mathrm{Ch}}=-\log\sum_{y'\in\mathcal Z}p_\theta(y'\mid x)\,\ind\{J(i,j,y')=u\},
\label{eq:channel}
\end{equation}
where $J(i,j,y')$ is the label the judge gives when the outcome is $y'$. In expectation $\ell_{\mathrm{Ch}}$ is minimized by $p_\theta=p^*$, but it needs the judge's labeling rule, so here it is a known-channel diagnostic, not a deployable remedy. $\ell_{\bt}$ needs no model of the judge; with the event as the response, it is the DPO loss~\cite{dpo} with $\beta=1$ and a constant reference. Its expectation, however, is minimized by a different distribution, the \emph{BT optimum}
\begin{equation}
b^*(\cdot\mid x)=\arg\min_{b}\;\E_{i,j\sim r,\;y\sim p^*}\big[\ell_{\bt}(b)\big],
\label{eq:bstar}
\end{equation}
which depends on $r$ and the judge as well as on $p^*$, but not on the model or its input. A candidate wins by being nearer to the realized outcome than its opponent, not by being probable, so an outcome with $p^*(z\mid x)=0$ can beat one that occurs; it then has positive probability under $b^*$. We measure this by the \emph{phantom mass},
\begin{equation}
\Phi_\theta(x)=\sum_{z:\,p^*(z\mid x)=0}p_\theta(z\mid x),
\label{eq:phantom}
\end{equation}
the probability a model places outside the support of $p^*$. The name makes no claim of impossibility: when $p^*$ is obtained by enumeration, a zero means that no enumerated case produces the outcome (\cref{sec:setup}). Total variation to $p^*$, which ignores where the zeros are, is reported beside it.

The three audits of \cref{fig:method} examine what such feedback does to a model. They are measurement procedures with stated information access, not a new backbone or training algorithm, and passing one does not imply passing another.

\subsection{Hazard probability readout}
\label{sec:readout}
The readout computes $q_\theta(x)$ in \cref{eq:hazard} from a documented event distribution, category identities, normalization rule and hazard subset. The rule matters because verbal confidence, sequence likelihood and normalized categorical probability can yield different quantities~\cite{know,askcal}. An exact readout sums the model's probability over every complete response that names each outcome, without sampling. When $p^*$ is known, the readout is scored against it on conditions unseen in training, and its change across states or actions is compared with the true change.

To isolate wording sensitivity, compare descriptions $x$ and $x'$ of one structure and action with $\theta$, $\mathcal Z$ and $\mathcal H$ fixed:
\begin{equation}
\Delta_{\mathrm{word}}(x,x')=100[q_\theta(x')-q_\theta(x)].
\label{eq:word}
\end{equation}
Descriptions must be semantically equivalent to qualify as a pair, and aggregation preserves the pairing~\cite{hallusion,semantic}.

\subsection{Pre-update response audit}
\label{sec:response}
The response audit asks what can be said about an update before its feedback is observed. It scores forecasts fixed in advance against the updates that follow, always beside competing forecasts, at two horizons.

\textbf{One update.}
Let $B$ denote a fresh feedback batch and $U(\theta,B)$ the update program. Its effect on the readout is $r_B(x)=q_{U(\theta,B)}(x)-q_\theta(x)$. For a gradient step, $U(\theta,B)=\theta-\eta g_B$ with $g_B=\nabla_\theta\ell_B(\theta)$, the standard first-order approximation is $d_B(x)=-\eta\nabla_\theta q_\theta(x)^\top g_B$; the reported fragility of such estimates motivates checks against finite updates~\cite{fragile}. Supplementary \S A.1 bounds $|r_B-d_B|$ under smoothness conditions that we do not verify numerically. The committed forecast $D^*$, in pp, targets the expectation of $d_B$ over batches and is fixed before any batch is drawn. Given independent responses with mean $\bar r$ and simultaneous half-width $h$, an operational check is
\begin{equation}
|100\bar r-D^*|+h\leq\epsilon_{\mathrm{op}},
\label{eq:operational}
\end{equation}
with all terms in pp. Passing it is not by itself evidence of predictive value, because a wrong forecast may pass too. Two comparisons decide that. The first is the \emph{zero forecast}, which predicts no change. For any committed forecast $f$,
\begin{equation}
\mathrm{RMSE}(f)^2=(100\bar r-f)^2+s^2,
\label{eq:floor}
\end{equation}
where $s$ is the root-mean-square spread of the per-batch responses about their mean. No committed forecast has an RMSE below $s$, and when $s$ exceeds $|100\bar r|$ the best one is little better than zero. The second is the per-batch diagnostic $L^*$, which evaluates $d_B$ with the observed batch. Its accuracy characterizes the local approximation after feedback arrives and cannot establish prospective performance.

\textbf{Converged endpoint.}
When training on a feedback dataset runs to a plateau, the natural forecast is the optimum of the objective's expected loss: $b^*$ for Event BT and $p^*$ for Full Channel, computed before training. We compare its total variation to the trained model with that of three competitors: the other objective's optimum, the pre-feedback model (no change), and a \emph{matched flattening}, the pre-feedback model mixed with the uniform distribution until its phantom mass equals that of $b^*$. The last tests whether $b^*$ predicts which outcomes gain probability, and not only how much leaves the reference support. Because $b^*$ uses $p^*$, this forecast is an audit instrument for settings with known outcome distributions, not a deployable predictor.

% Figure 3 (the push task of Sec. 4.1) spans both columns. It is declared here, on page 4, so that it is set at
% the top of page 5, where Sec. 4.1 begins.
\begin{figure*}[t]
\centering
\includegraphics[width=\textwidth]{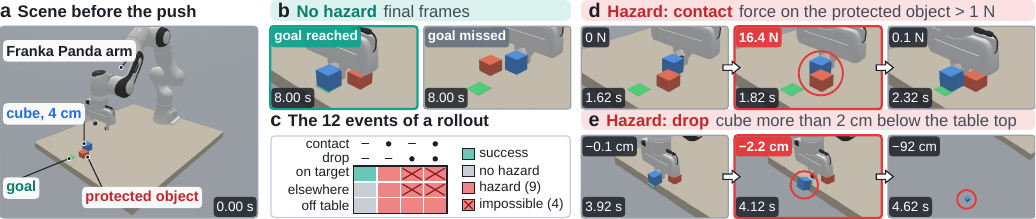}
\caption{Push task in simulation; frames from four different branches. (a)~Initial state. (b)~Final frames without a hazard. (c)~The 12 events of a rollout: final region, contact and drop. (d,\,e)~The two hazards 0.2~s before, at (red outline) and 0.5~s after their first recorded frame, with the contact force and the cube's lowest point relative to the table top. The model sees only text: layout, stated gap and push.}
\label{fig:task}
\vspace{\taskVS}
\end{figure*}

\subsection{Fixed-threshold decision audit}
\label{sec:decision}
The decision audit asks what the probabilities do to actions under a common rule. The model executes when $q_\theta(x)\leq\tau$; the oracle applies the same rule to $q^*(x)$. A false allow (FA) executes where the oracle rejects; a false reject (FR) rejects where the oracle executes. Costs enter in one of two forms. In the reference binary decision, executing costs $q^*(x)$, a unit cost on a hazard, and rejecting costs the constant $\tau$, so that per-context regret is
\begin{equation}
\mathcal R_\theta(x)=|q^*(x)-\tau|\,
\ind\{a_\theta(x)\ne a^*(x)\},
\label{eq:regret}
\end{equation}
with $a_\theta$ and $a^*$ the model and oracle actions. With an explicit success outcome $z_0$, executing has utility
\begin{equation}
u_\lambda(x)=p^*(z_0\mid x)-\lambda\,q^*(x)
\label{eq:utility}
\end{equation}
and rejecting has utility 0, where $\lambda$ is the cost of a hazard relative to the value of a success. In both forms an error that crosses the boundary changes the decision while a larger error on the same side may not, so average probability error does not determine regret (Supplementary \S A.2).

Two threshold regimes separate two kinds of damage. With an \emph{inherited} threshold, $\tau$ is chosen for the pre-feedback model and kept for every readout. This is the primary regime, because feedback is used precisely when ground truth for re-tuning is scarce. With a \emph{re-tuned} threshold, $\tau$ is chosen for each readout on held-out truth. A readout that inflates $q_\theta$ but preserves its order across conditions loses under the first regime and not under the second: a calibration loss, as opposed to a ranking loss. \emph{Action choice} removes the threshold altogether. Among the candidate actions of one layout and rejection, the model picks the highest predicted utility, and its regret is measured against the oracle's pick.

\textbf{Information access.}
Paired descriptions are fixed before wording sensitivity is measured, and forecasts are committed before the updates they are scored against. Decisions use a common rule, with every declared threshold and cost ratio reported; oracle truth is reserved for evaluation, and readouts trained with additional truth information are marked (*).

\section{Experiments}
\label{sec:experiments}
\subsection{Setup: A simulated push}
\label{sec:setup}
A Franka arm in ManiSkill~\cite{maniskill3} pushes a 4~cm cube toward a goal past a protected object beside its path (\cref{fig:task}). A condition $x$ is a layout with one of three pushes fixed in advance (two speeds and a short push), given to the model as text. Each rollout ends in one of $K=12$ events, the combinations of final region (target, elsewhere on the table, off the table), contact with the protected object and drop; the nine events with a drop or a contact are hazards. The outcome depends on quantities the model never sees, namely friction (four values), control latency (four) and an error of up to $\pm2$~mm in the stated gap (five). The environment draws them uniformly from this grid, so the reference $p^*(\cdot\mid x)$ is the exact outcome frequency over 80 enumerated branches.

\textbf{What a zero of $p^*$ means.}
An event with $p^*(z\mid x)=0$ is produced by none of the 80 branches of $x$: a statement about the enumerated environment, not about physical possibility. Two kinds of zero differ. Four of the twelve events pair a drop below the table top with a final position on the table (crossed in \cref{fig:task}c); no hidden parameters produce them, and we call them \emph{impossible}. Every other zero is merely absent from the condition's 80 branches, and a denser grid could give it positive probability.

% Table 1 (main results) is declared here, so that it is set at the top of page 6, with Sec. 4.2.
% written by tools/make_main_tables.py; do not edit
\begin{table*}[t]
\caption{Main results at $N=64$. (a)~$\Phi$ in pp on the 153 fractional trained conditions; other metrics against $p^*$ on all 213 (KL, entropy in nats; $p^*$ has entropy 0.526; TV: total variation; Mode: share of conditions with $p^*$'s most likely event; ECE: expected calibration error over events $z$ or of $q$; MAE: mean absolute error of $q$); utility in pp per decision at $\lambda=3$ (inherited $\tau=0.23$); false rejects (FR) and allows (FA) in \%; *: additional truth information. (b)~In pp (KL in nats); a row changes one property of the main round (12 events, 80 branches, trained pool, $w=0.2$ in $r=(1-w)\,p_{\theta_0}+w/12$). Bold (a): Event BT worst of all readouts (columns with arrows). Gray (b): the 95\% interval includes zero or favors Event BT. Icons as in \cref{fig:method}. $^\dagger$Pooled over six seeds (KL, Brier: range). $^\ddagger$$t$ interval over seed means includes zero.}
\label{tab:main}
\centering\footnotesize
\renewcommand{\arraystretch}{1.04}\setlength{\tabrowht}{1.04\baselineskip}
\setlength{\aboverulesep}{2pt}\setlength{\belowrulesep}{2.4pt}
\tabmeasure{A1}{{Readout},{\rmark{frozen}Frozen ($\theta_0$)\ricon{\includegraphics[height=7pt]{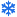}}},{\rmark{bt}Event BT\ricon{\includegraphics[height=7pt]{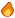}}},{\rmark{fc}Full Channel\ricon{\includegraphics[height=7pt]{figs/icons/flame.pdf}}},{\rmark{mle}Outcome MLE*\ricon{\includegraphics[height=7pt]{figs/icons/flame.pdf}}},{\rmark{self}Self control\ricon{\includegraphics[height=7pt]{figs/icons/flame.pdf}}}}
\tabmeasure{A2}{{Seed},{\na},{1},{2},{1},{2},{1},{1}}
\tabmeasure{A3}{{$\Phi$\,$\downarrow$},{0.8},{\tabsig{39.0}},{\tabsig{43.1}},{0.5},{0.6},{0.4},{1.1}}
\tabmeasure{A4}{{KL\,$\downarrow$},{0.025},{\tabsig{0.430}},{\tabsig{0.476}},{0.029},{0.030},{0.016},{0.038}}
\tabmeasure{A5}{{Brier\,$\downarrow$},{0.007},{\tabsig{0.109}},{\tabsig{0.122}},{0.011},{0.012},{0.006},{0.013}}
\tabmeasure{A6}{{TV\,$\downarrow$},{0.049},{\tabsig{0.324}},{\tabsig{0.346}},{0.061},{0.061},{0.045},{0.065}}
\tabmeasure{A7}{{Entropy},{0.539},{1.425},{1.449},{0.517},{0.534},{0.535},{0.558}}
\tabmeasure{A8}{{Mode\,$\uparrow$},{0.954},{0.949},{0.939},{0.942},{0.907},{0.945},{0.935}}
\tabmeasure{A9}{{ECE$_z$\,$\downarrow$},{0.002},{\tabsig{0.040}},{\tabsig{0.036}},{0.003},{0.002},{0.002},{0.002}}
\tabmeasure{A10}{{ECE$_q$\,$\downarrow$},{0.015},{\tabsig{0.077}},{\tabsig{0.095}},{0.027},{0.019},{0.015},{0.022}}
\tabmeasure{A11}{{Utility\,$\uparrow$},{13.21},{\tabsig{10.26}},{\tabsig{10.28}},{13.44},{13.44},{13.52},{12.03}}
\tabmeasure{A12}{{FR\,$\downarrow$},{1.7},{\tabsig{12.8}},{\tabsig{13.2}},{1.7},{1.5},{2.0},{5.9}}
\tabmeasure{A13}{{FA\,$\downarrow$},{3.7},{0.0},{0.0},{4.0},{3.1},{2.6},{3.3}}
\tabmeasure{A14}{{Utility\,$\uparrow$},{13.21},{13.37},{13.06},{12.65},{13.31},{13.43},{13.47}}
\tabmeasure{A15}{{$\tau$},{0.23},{0.46},{0.42},{0.18},{0.22},{0.25},{0.31}}
\tabgrid{A}{10,20,10,10,10,10,10,10,10,20,10,10,20,10}
\begin{tabular}{@{\hspace{\tabpadx}}w{l}{\tabw{A1}}@{\hspace{\tabg{A1}}}w{c}{\tabw{A2}}@{\hspace{\tabg{A2}}}w{l}{\tabw{A3}}@{\hspace{\tabg{A3}}}w{r}{\tabw{A4}}@{\hspace{\tabg{A4}}}w{r}{\tabw{A5}}@{\hspace{\tabg{A5}}}w{r}{\tabw{A6}}@{\hspace{\tabg{A6}}}w{r}{\tabw{A7}}@{\hspace{\tabg{A7}}}w{r}{\tabw{A8}}@{\hspace{\tabg{A8}}}w{r}{\tabw{A9}}@{\hspace{\tabg{A9}}}w{r}{\tabw{A10}}@{\hspace{\tabg{A10}}}w{r}{\tabw{A11}}@{\hspace{\tabg{A11}}}w{l}{\tabw{A12}}@{\hspace{\tabg{A12}}}w{r}{\tabw{A13}}@{\hspace{\tabg{A13}}}w{r}{\tabw{A14}}@{\hspace{\tabg{A14}}}w{r}{\tabw{A15}}@{\hspace{\tabpadx}}}
\toprule
\multicolumn{2}{@{\hspace{\tabpadx}}l@{}}{\rlap{\tblock{a}{Main round}}} & \multicolumn{8}{@{}l@{}}{\tabbanner{A}{3}{10}{\includegraphics[height=7.6pt]{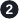}}{Response}{the predicted distribution, against $p^*$}} & \multicolumn{5}{@{}l@{\hspace{\tabpadx}}}{\tabbannersplit{A}{11}{14}{15}{\includegraphics[height=7.6pt]{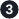}}{Decision}{inherited $\tau$}{re-tuned $\tau$}}\\[1.2pt]
Readout & Seed & $\Phi$\,$\downarrow$ & KL\,$\downarrow$ & Brier\,$\downarrow$ & TV\,$\downarrow$ & Entropy & Mode\,$\uparrow$ & ECE$_z$\,$\downarrow$ & ECE$_q$\,$\downarrow$ & Utility\,$\uparrow$ & FR\,$\downarrow$ & FA\,$\downarrow$ & Utility\,$\uparrow$ & $\tau$\\
\midrule
\rmark{frozen}Frozen ($\theta_0$)\ricon{\includegraphics[height=7pt]{figs/icons/snowflake.pdf}} & \na & 0.8 & 0.025 & 0.007 & 0.049 & 0.539 & 0.954 & 0.002 & 0.015 & 13.21 & 1.7 & 3.7 & 13.21 & 0.23\\
\addlinespace[1.6pt]
\rmark{bt}Event BT\ricon{\includegraphics[height=7pt]{figs/icons/flame.pdf}} & 1 & \tabsig{39.0} & \tabsig{0.430} & \tabsig{0.109} & \tabsig{0.324} & 1.425 & 0.949 & \tabsig{0.040} & \tabsig{0.077} & \tabsig{10.26} & \tabsig{12.8} & 0.0 & 13.37 & 0.46\\
 & 2 & \tabsig{43.1} & \tabsig{0.476} & \tabsig{0.122} & \tabsig{0.346} & 1.449 & 0.939 & \tabsig{0.036} & \tabsig{0.095} & \tabsig{10.28} & \tabsig{13.2} & 0.0 & 13.06 & 0.42\\
\addlinespace[1.6pt]
\rmark{fc}Full Channel\ricon{\includegraphics[height=7pt]{figs/icons/flame.pdf}} & 1 & 0.5 & 0.029 & 0.011 & 0.061 & 0.517 & 0.942 & 0.003 & 0.027 & 13.44 & 1.7 & 4.0 & 12.65 & 0.18\\
 & 2 & 0.6 & 0.030 & 0.012 & 0.061 & 0.534 & 0.907 & 0.002 & 0.019 & 13.44 & 1.5 & 3.1 & 13.31 & 0.22\\
\addlinespace[1.6pt]
\rmark{mle}Outcome MLE*\ricon{\includegraphics[height=7pt]{figs/icons/flame.pdf}} & 1 & 0.4 & 0.016 & 0.006 & 0.045 & 0.535 & 0.945 & 0.002 & 0.015 & 13.52 & 2.0 & 2.6 & 13.43 & 0.25\\
\rmark{self}Self control\ricon{\includegraphics[height=7pt]{figs/icons/flame.pdf}} & 1 & 1.1 & 0.038 & 0.013 & 0.065 & 0.558 & 0.935 & 0.002 & 0.022 & 12.03 & 5.9 & 3.3 & 13.47 & 0.31\\
\bottomrule
\end{tabular}
\\[6pt]
\tabsetcol{a}{{+38.5},{+42.5},{+29.1},{+26.6},{+33.7},{+36.4},{+39.3},{+41.7},{+33.8},{+40.1},{+38.3},{+38.1},{+37.8},{+37.7},{+38.2},{+36.5}}{{+29.0},{+29.5}}
\tabsetcol{b}{{+0.40},{+0.45},{+0.28},{+0.25},{+0.34},{+0.41},{+0.28},{+0.31},{+0.35},{+0.45},{+0.39},{+0.40},{+0.39},{+0.40},{+0.40},{+0.37}}{{+0.25--0.32},{+0.19--0.22}}
\tabsetcol{c}{{+0.097},{+0.110},{+0.059},{+0.049},{+0.078},{+0.106},{+0.068},{+0.074},{+0.078},{+0.110},{+0.085},{+0.094},{+0.086},{+0.090},{+0.098},{+0.090}}{{+0.049--0.081},{+0.038--0.053}}
\tabsetcol{d}{{+5.5},{+7.4},{$-$0.3},{$-$0.5},{+6.9},{+11.6},{+5.9},{+7.3},{+10.9},{+10.5},{+4.4},{+4.0},{+2.1},{+3.5},{+10.6},{+6.2}}{{+0.76$^\ddagger$},{+0.65$^\ddagger$}}
\tabsetcol{e}{{$-$3.2},{$-$3.2},{+0.8},{$-$1.1},{$-$2.8},{$-$4.5},{$-$1.3},{$-$1.8},{$-$6.9},{$-$6.3},{$-$1.8},{$-$2.5},{$-$1.9},{$-$2.6},{$-$2.7},{$-$2.3}}{{$-$0.60$^\ddagger$},{$-$0.39$^\ddagger$}}
\tabsetcol{f}{{+4.7},{+5.9},{+1.1},{+1.2},{+3.3},{+7.3},{+4.7},{+3.8},{+7.6},{+7.6},{+2.7},{+3.0},{+3.0},{+3.0},{+3.0},{+2.3}}{{+1.6},{+2.8}}
\tabmeasure{B1}{{\tabgroup{Event space}},{\tabgroup{Reference}},{\tabgroup{Pool}},{\tabgroup{Candidate law}}}
\tabmeasure{B2}{{8 feasible events},{8 feasible events, six seeds$^\dagger$},{252 branches},{unseen},{unseen, 8 feasible events, six seeds$^\dagger$},{$w=0.05$},{$w=0.5$},{uniform ($w=1$)},{centered on $p^*$}}
\tabmeasure{B3}{{\hbox to \tabpairw{a}{}},{$\Delta\Phi$}}
\tabmeasure{B4}{{\hbox to \tabpairw{b}{}},{$\Delta$KL}}
\tabmeasure{B5}{{\hbox to \tabpairw{c}{}},{$\Delta$Brier}}
\tabmeasure{B6}{{\hbox to \tabpairw{d}{}},{$\Delta$MAE$_q$}}
\tabmeasure{B7}{{\hbox to \tabpairw{e}{}},{$\Delta$Utility}}
\tabmeasure{B8}{{\hbox to \tabpairw{f}{}},{$\Delta$Regret}}
\tabgrid{B}{8,20,10,10,20,20,10}
\begin{tabular}{@{\hspace{\tabpadx}}w{l}{\tabw{B1}}@{\hspace{\tabg{B1}}}w{l}{\tabw{B2}}@{\hspace{\tabg{B2}}}w{c}{\tabw{B3}}@{\hspace{\tabg{B3}}}w{c}{\tabw{B4}}@{\hspace{\tabg{B4}}}w{c}{\tabw{B5}}@{\hspace{\tabg{B5}}}w{c}{\tabw{B6}}@{\hspace{\tabg{B6}}}w{c}{\tabw{B7}}@{\hspace{\tabg{B7}}}w{c}{\tabw{B8}}@{\hspace{\tabpadx}}}
\toprule
\multicolumn{2}{@{\hspace{\tabpadx}}l@{}}{\rlap{\tblock{b}{\rmark{bt}Event BT $-$ \rmark{fc}Full Channel}}} & \multicolumn{3}{@{}l@{}}{\tabbanner{B}{3}{5}{\includegraphics[height=7.6pt]{figs/icons/badge2.pdf}}{Response}{the predicted distribution}} & \multicolumn{1}{@{}l@{}}{\tabbanner{B}{6}{6}{\includegraphics[height=7.6pt]{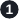}}{Readout}{}} & \multicolumn{2}{@{}l@{\hspace{\tabpadx}}}{\tabbanner{B}{7}{8}{\includegraphics[height=7.6pt]{figs/icons/badge3.pdf}}{Decision}{inherited $\tau$}}\\[1.2pt]
\multicolumn{2}{@{\hspace{\tabpadx}}l@{\hspace{\tabg{B2}}}}{Setting\hfill\tabns{each cell: seed 1, seed 2}} & $\Delta\Phi$ & $\Delta$KL & $\Delta$Brier & $\Delta$MAE$_q$ & $\Delta$Utility & $\Delta$Regret\\
\midrule
\multicolumn{2}{@{\hspace{\tabpadx}}l@{\hspace{\tabg{B2}}}}{\textbf{Main round}} & \tabpair{a}{1}{+38.5}{1}{+42.5} & \tabpair{b}{1}{+0.40}{1}{+0.45} & \tabpair{c}{1}{+0.097}{1}{+0.110} & \tabpair{d}{1}{+5.5}{1}{+7.4} & \tabpair{e}{1}{$-$3.2}{1}{$-$3.2} & \tabpair{f}{1}{+4.7}{1}{+5.9}\\
\addlinespace[2.4pt]
\tabgroup{Event space} & 8 feasible events & \tabpair{a}{1}{+29.1}{1}{+26.6} & \tabpair{b}{1}{+0.28}{1}{+0.25} & \tabpair{c}{1}{+0.059}{1}{+0.049} & \tabpair{d}{0}{$-$0.3}{0}{$-$0.5} & \tabpair{e}{0}{+0.8}{1}{$-$1.1} & \tabpair{f}{0}{+1.1}{0}{+1.2}\\
 & 8 feasible events, six seeds$^\dagger$ & \tabwide{a}{1}{+29.0} & \tabwide{b}{1}{+0.25--0.32} & \tabwide{c}{1}{+0.049--0.081} & \tabwide{d}{1}{+0.76$^\ddagger$} & \tabwide{e}{1}{$-$0.60$^\ddagger$} & \tabwide{f}{1}{+1.6}\\
\addlinespace[2.4pt]
\tabgroup{Reference} & 252 branches & \tabpair{a}{1}{+33.7}{1}{+36.4} & \tabpair{b}{1}{+0.34}{1}{+0.41} & \tabpair{c}{1}{+0.078}{1}{+0.106} & \tabpair{d}{1}{+6.9}{1}{+11.6} & \tabpair{e}{1}{$-$2.8}{1}{$-$4.5} & \tabpair{f}{0}{+3.3}{1}{+7.3}\\
\addlinespace[2.4pt]
\tabgroup{Pool} & unseen & \tabpair{a}{1}{+39.3}{1}{+41.7} & \tabpair{b}{1}{+0.28}{1}{+0.31} & \tabpair{c}{1}{+0.068}{1}{+0.074} & \tabpair{d}{1}{+5.9}{1}{+7.3} & \tabpair{e}{1}{$-$1.3}{1}{$-$1.8} & \tabpair{f}{1}{+4.7}{0}{+3.8}\\
 & unseen, 8 feasible events, six seeds$^\dagger$ & \tabwide{a}{1}{+29.5} & \tabwide{b}{1}{+0.19--0.22} & \tabwide{c}{1}{+0.038--0.053} & \tabwide{d}{1}{+0.65$^\ddagger$} & \tabwide{e}{1}{$-$0.39$^\ddagger$} & \tabwide{f}{1}{+2.8}\\
\addlinespace[2.4pt]
\tabgroup{Candidate law} & $w=0.05$ & \tabpair{a}{1}{+33.8}{1}{+40.1} & \tabpair{b}{1}{+0.35}{1}{+0.45} & \tabpair{c}{1}{+0.078}{1}{+0.110} & \tabpair{d}{1}{+10.9}{1}{+10.5} & \tabpair{e}{1}{$-$6.9}{1}{$-$6.3} & \tabpair{f}{1}{+7.6}{1}{+7.6}\\
 & $w=0.5$ & \tabpair{a}{1}{+38.3}{1}{+38.1} & \tabpair{b}{1}{+0.39}{1}{+0.40} & \tabpair{c}{1}{+0.085}{1}{+0.094} & \tabpair{d}{1}{+4.4}{1}{+4.0} & \tabpair{e}{1}{$-$1.8}{1}{$-$2.5} & \tabpair{f}{1}{+2.7}{1}{+3.0}\\
 & uniform ($w=1$) & \tabpair{a}{1}{+37.8}{1}{+37.7} & \tabpair{b}{1}{+0.39}{1}{+0.40} & \tabpair{c}{1}{+0.086}{1}{+0.090} & \tabpair{d}{1}{+2.1}{1}{+3.5} & \tabpair{e}{1}{$-$1.9}{1}{$-$2.6} & \tabpair{f}{1}{+3.0}{1}{+3.0}\\
 & centered on $p^*$ & \tabpair{a}{1}{+38.2}{1}{+36.5} & \tabpair{b}{1}{+0.40}{1}{+0.37} & \tabpair{c}{1}{+0.098}{1}{+0.090} & \tabpair{d}{1}{+10.6}{1}{+6.2} & \tabpair{e}{1}{$-$2.7}{1}{$-$2.3} & \tabpair{f}{1}{+3.0}{0}{+2.3}\\
\bottomrule
\end{tabular}
\vspace{\tableVS}
\end{table*}

\textbf{Model and feedback.}
The world model is Qwen2.5-1.5B-Instruct~\cite{qwen25} with a rank-8 LoRA adapter~\cite{lora}, trained on 399 conditions to minimize the expected negative log-likelihood under $p^*$. Its distribution over the 12 events is read exactly from the probabilities of 24 complete response paths, two per event; every objective and audit uses this distribution. Feedback follows \cref{sec:formulation}: candidates are drawn from $r=0.8\,p_{\theta_0}+0.2/12$, and the judge's distance counts a disagreement in region or drop once and in contact twice. Event BT and Full Channel receive identical comparisons. From the same starting model (Frozen, $\theta_0$) we train Event BT, Full Channel, Outcome MLE* (which sees the outcome $y$ itself, additional truth information) and a \emph{self control}: Full Channel on outcomes drawn from $\theta_0$'s own prediction, whose optimum is $\theta_0$, so that any change is drift of the training itself. Each uses $N\in\{16,64,256,1024\}$ comparisons (or outcome draws) per condition; $N=64$ was fixed in advance as the realistic budget, and $N=64$ and 1024 have two feedback seeds. Event BT and Full Channel are also trained on exact expected feedback ($N=\infty$). This gives 26 runs with one recipe, each stopped at a plateau of its held-out loss.

\textbf{Evaluation.}
Training and the primary evaluation use 213 conditions that $\theta_0$ already fits (hazard MAE 2.0~pp), so a truth-seeking objective has nothing to correct and movement is attributable to the objective; 153 of them are \emph{fractional} ($0<q^*<1$). A pool of 300 conditions never used in training tests transfer. Beside $\Phi$ and the hazard error we report KL divergence and Brier score against $p^*$, utility under an inherited or re-tuned threshold, and the regret of action choice (\cref{sec:decision}). Pass criteria, analysis code, feedback data and forecasts were frozen before the runs, and intervals are 95\% bootstrap intervals over layout families. The supplementary material presents the full protocol, a superseded single-dose round, all deviations and the real-robot experiments.

\begin{figure*}[t]
\centering
\includegraphics[width=\linewidth]{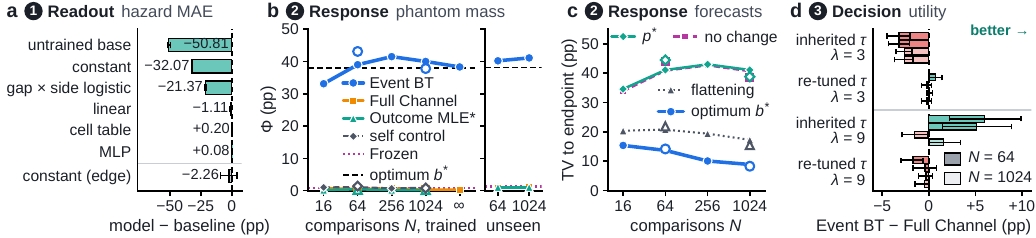}
\caption{Push study: one starting model through the three audits. (a)~Hazard MAE on blind layouts, model minus baseline; negative values favor the model, whose own MAE is 2.4~pp (gray: layouts that vary the clearance to the table edge). (b)~Phantom mass on the 153 fractional trained conditions and on the unseen pool; $N=\infty$ is training on expected feedback. (c)~Total variation (TV) from the Event BT endpoint to forecasts fixed before training. (d)~Utility per decision on the 213 trained conditions; the two bars of a budget are the two feedback seeds (red: Event BT worse; green: better). Whiskers: 95\% intervals; open markers: second feedback seed; *: additional truth information.}
\label{fig:push}
\vspace{\pushVS}
\end{figure*}

\textbf{The starting model is accurate.}
On 19 blind layout families (588 contact conditions), whose labels were opened only after the predictions were frozen, hazard MAE is 2.4~pp (4.9~pp on fractional conditions) and $q_\theta$ correlates with $q^*$ at $+0.994$. Its error in the change of $q$ is 4.1~pp across gaps and 4.9~pp across pushes, under half that of a zero-response baseline. It is level with a 32-unit MLP on the same geometry features ($+0.08$~pp, interval $[-0.27,+0.47]$; \cref{fig:push}a): the language model adds nothing that a small numeric model lacks, and the study needs nothing more than an accurate model for feedback to act on. On layouts varying the clearance to the table edge it fails (MAE 29.0~pp against 31.3 for a constant), so every result below concerns contact layouts.

\subsection{Main results}
\label{sec:exp-main}
\Cref{tab:main} gives every readout on the main round (a) and the contrast Event BT minus Full Channel in ten settings (b).

\begin{figure}[t]
\centering
\includegraphics[width=\linewidth]{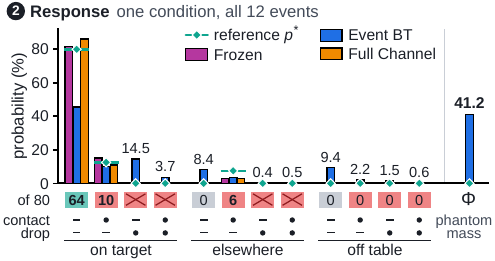}
\caption{Push study, one trained condition ($N=64$): event probabilities and, right, phantom mass $\Phi$. Tiles as in \cref{fig:task}c, with how many of the 80 branches end in the event (crossed: impossible); numbers: Event BT where $p^*=0$.}
\label{fig:events}
\vspace{\eventsVS}
\end{figure}

\textbf{Event BT moves an accurate model; the other objectives do not.}
From the same starting model and the same comparisons, Event BT raises the phantom mass from 0.8 to 39.0 and 43.1~pp, while Full Channel, Outcome MLE* and the self control stay at or below 1.4~pp. The whole distribution suffers: KL divergence from $p^*$ grows from 0.025 to 0.430 and 0.476~nats, against 0.029 and 0.030 for Full Channel. Event BT does not lose the truth: its most likely event agrees with that of $p^*$ nearly as often as Frozen's, and it misses almost none of the reference mass. It adds probability to events that do not occur, so the entropy of its predictions is 1.425 and 1.449~nats against 0.526 for $p^*$. \Cref{fig:events} shows one condition, picked by a rule fixed before its data were seen (Supplementary \S C.7): on the nine events its 80 branches never produce, Frozen and Full Channel leave under 0.01~pp and Event BT 41.2~pp, 19.1~pp on impossible events, so its hazard probability is 36.7\% against a true 20.0\%.

\textbf{The shift appears in every setting.}
Event BT's phantom mass exceeds Full Channel's by 26.6 to 42.5~pp, and its KL divergence and Brier score are worse, in all ten settings and with every interval excluding zero: on the eight feasible events, against a denser reference, on unseen conditions, over six feedback seeds and under five candidate laws.

\textbf{Its cost depends on where the probability lands.}
With all 12 events, part of the phantom mass falls on impossible events that count as hazards. Hazard error is then 2.1 to 11.6~pp above Full Channel's, and a threshold inherited from Frozen loses utility in every such setting: 3.2~pp per decision on the trained conditions, 1.3 and 1.8~pp on unseen ones, and 1.8 to 6.9~pp under the other candidate laws. On the eight feasible events the pooled utility loss stays below the 1~pp margin fixed in advance, while the regret of action choice remains higher, by 1.6~pp on the trained and 2.8~pp on the unseen conditions.

\textbf{Re-tuning the threshold removes the utility loss.}
With a threshold re-tuned for each readout, Event BT is within 1~pp of Full Channel in every setting: the feedback damages calibration, not the ordering of conditions by risk. But re-tuning takes the held-out truth that comparative feedback is meant to replace, and the distribution itself stays wrong.

% Figure 4 is declared here, so that it is set at the top of page 7, with the analyses that read it.

\subsection{Response: Feedback moves the model\texorpdfstring{\\}{ }to the objective's optimum}
\label{sec:exp-response}
\textbf{The shift.}
Event BT's phantom mass is 33.1~pp already at $N=16$ and between 37.8 and 43.1~pp from $N=64$ (\cref{fig:push}b), near the level of $b^*$ itself (37.9~pp) and of training on expected feedback (38.3~pp). It exceeds the self control's drift by 32.2 to 41.7~pp at every budget and seed, and its endpoint is 19.2 to 33.0~pp closer in total variation to $b^*$ than to $p^*$. The shift is thus specific to the objective, not to continued training, and more comparisons do not average it away: it is where the objective's optimum lies.

\textbf{The optimum forecasts the endpoint.}
The test is whether finite data and early stopping reach the minimizer, and whether $b^*$ says where probability goes. $b^*$ was computed (Supplementary \S A.3) and hashed before training. It lies 15.4~pp in total \mbox{variation} from the Event BT endpoint at $N=16$ and 8.3 to 8.9~pp at $N=1024$, against 34 to 45~pp for $p^*$ and the unchanged model (\cref{fig:push}c). It is also 5.0 to 9.4~pp closer than the matched flattening, with intervals excluding zero at every budget, so it predicts which outcomes outside the support gain probability, not only how much. For Full Channel, $p^*$ is the better forecast, by 31.8 to 35.8~pp, though not separable from no change at $N\leq64$.

\textbf{One update cannot be forecast.}
With the forecast $D^*$ committed first and 32 fresh batches per objective, Event BT's mean response at the primary step size is $+1.131$~pp with a batch spread of 4.148~pp, against $D^*=+0.213$~pp (Supplementary \S C.8); by \cref{eq:floor} no committed forecast could beat zero by more than about 4\% (derived). The accuracy check passes, but it also passes when Event BT's forecast is scored against Full Channel's responses, and the non-triviality test, which requires the interval of the realized change to exclude zero, is inconclusive for every target and step size. Given the batch, the same first-order calculation is nearly exact (median error 0.0010 and 0.0030~pp): post-hoc accuracy says little about what can be committed in advance.

\subsection{Decision: False rejects\texorpdfstring{\\}{ }under an inherited threshold}
\label{sec:exp-decision}
The threshold is chosen for Frozen on three reserve layout families ($\tau=0.23$ at the primary cost ratio $\lambda=3$) and kept for every readout. Under it Event BT's false rejects rise to 12.8 and 13.2\% of the trained conditions, against 1.7\% for Frozen, and it makes no false allow (\cref{tab:main}a). The comparison fixed in advance, Event BT minus Full Channel, is $-3.2$~pp of utility in both seeds (intervals $[-4.5,-2.0]$ and $[-4.3,-2.1]$) and $-2.6$~pp at $N=1024$ (\cref{fig:push}d); $\lambda=1$ gives the same picture, and on the 300 unseen conditions the criterion fixed in advance for a loss is met. \mbox{Re-tuning} each readout's threshold on the reserve families removes it at every budget and cost ratio (differences between $-1.7$ and $+0.7$~pp), as Event BT's thresholds rise to 0.42 to 0.46.

Three observations qualify this picture. First, action choice uses no threshold, yet Event BT's regret exceeds Full Channel's by 4.7~pp $[1.1,9.1]$ and 5.9~pp $[0.9,12.5]$ at $N=64$: phantom mass is not spread evenly over the pushes of a layout. Second, the sign depends on the cost ratio: at $\lambda=9$ and $N=64$, Event BT is 5.1 to 6.0~pp \emph{better} than Full Channel, whose 7.1 and 7.5\% false allows are expensive there, although both fall below Frozen. The inflated hazard probability acts as an accidental margin, and the advantage does not persist at $N=1024$ ($-1.5$ and $+1.6$~pp). Third, the self control, which has nothing to learn, loses 1.2 and 1.6~pp under the inherited threshold (derived), so part of the loss against Frozen comes from continued training on a finite sample and not from the objective. The contrast fixed in advance, with Full Channel, shares recipe and comparisons.

\subsection{Robustness: Event space, reference\texorpdfstring{\\}{ }and candidate law}
\label{sec:exp-robust}
The reference could produce these results by itself, through its four impossible events or its coarse grid, on which a rare event counts as one that never occurs; so could the candidate law. Three experiments change one of them each; their criteria were fixed in advance (Supplementary \S C.9, \S C.10).

\begin{figure}[t]
\centering
\includegraphics[width=\linewidth]{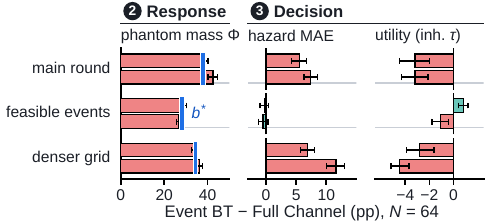}
\caption{Push study, robustness at $N=64$: Event BT minus Full Channel with 95\% intervals, in the main round, on the eight feasible events and on a 252-branch reference (the two bars of a setting are the two feedback seeds; blue: $\Phi$ of each optimum $b^*$). Utility is under the inherited threshold. Red: Event BT worse; green: better.}
\label{fig:robust}
\vspace{\robustVS}
\end{figure}

\textbf{Feasible events only.}
Of Event BT's 39.0 and 43.1~pp of phantom mass, 17.1 and 20.6~pp lie on the impossible events, all of them hazards. Restricted to the eight feasible events and with all arms retrained, 30.1 and 27.4~pp remain on feasible events that the condition never produces, the level of the restricted optimum (28.2~pp; \cref{fig:robust}); pooled over six feedback seeds, Event BT exceeds Full Channel by 29.0~pp $[28.0,30.0]$. The decision effects shrink to the edge of detection. Hazard error is higher by 0.76~pp $[0.34,1.16]$ and utility under the inherited threshold lower by 0.60~pp $[0.32,0.86]$, each with the harmful sign in four of six seeds and an interval over seeds that includes zero; the 1~pp margin fixed in advance is not met. Regret is higher by 1.61~pp $[0.24,3.21]$, in all six seeds. At $N=1024$ a gap returns (hazard MAE $+3.0$~pp, utility $-1.8$~pp, one seed).

% Figure 5 (cited in Sec. 4.2) is declared here, late in Sec. 4.5, so that it is set at the top of the left column of page 8, beside Figure 6.

\textbf{A denser reference.}
A nested grid of 252 branches per condition moves the reference by 4.0~pp of total variation on average. Retrained on it, Event BT has $\Phi=34.0$ and 36.9~pp, again at its optimum (34.4~pp, down from 37.9), and the observability and forecast criteria pass. The stability margin fixed in advance does not: $\Phi$ falls by 4.9 and 6.2~pp against the main-round runs, with 90\% intervals that reach beyond $-5$~pp, although scoring the main-round models against the denser reference changes $\Phi$ by only $-1.4$~pp. The decision loss remains, as the impossible events are back: $-2.8$ and $-4.5$~pp of utility under the inherited threshold.

\textbf{The candidate law.}
Both objectives were retrained on feedback drawn under four other laws: a uniform share $w$ of 0.05, 0.5 or 1 in $r=(1-w)\,p_{\theta_0}+w/12$, and candidates centered on $p^*$ instead of $p_{\theta_0}$. The optimum's phantom mass hardly changes (37.8 to 39.1~pp), Event BT ends at 34.6 to 41.1~pp against 0.3 to 1.0~pp for Full Channel, and every criterion fixed in advance passes. The consequences depend on the law: the utility loss is largest when candidates concentrate near the model's prediction ($-6.9$ and $-6.3$~pp at $w=0.05$), smaller for $w\geq0.5$ ($-1.8$ to $-2.6$~pp), and present even for candidates centered on the truth.

\subsection{Supporting evidence: Synthetic structures}
\label{sec:exp-structures}
Four earlier studies on synthetic finite-event environments, each described in text, first used the audits; with incomplete records and exposed structures (Supplementary \S D), they serve as illustrations, not as independent evidence. With weights frozen, reordering a description changes the hazard probability by a mean of 2.83~pp across eight pairs (maximum 15.56~pp; Supp.~Fig.~S1). A single update is no easier to forecast there: with 32 paired batches at each of two learning rates, all six mean-response bounds meet the 0.10~pp tolerance, but so do four of the six when zero replaces $D^*$ (derived; Supp.~Fig.~S2). Small differences change decisions: in mean error, Event BT is 0.14~pp below Frozen and Full Channel 0.27~pp above, yet on 12 exposed structures each differs from Frozen by one false reject at one of three thresholds, and on ten new wordings all three readouts make a false allow not seen on the matched originals (Supp.~Fig.~S3).

% Figure 6 is declared here, in the last subsection, so that it is set at the top of the right column of page 8, beside Figure 5.

\section{Conclusion}
Comparative feedback does not by itself teach a language world model the true outcome probabilities. On a simulated push, a Bradley--Terry objective with feedback from a truthful judge pulls an initially accurate model to that objective's optimum: 27 to 43~pp of probability on outcomes that a condition never produces, across event spaces, reference grids and candidate laws, on trained and unseen conditions. A likelihood that knows the judge's rule uses the same comparisons without the shift, so the shift comes from the objective. Its cost depends on where the misplaced probability lands: on impossible hazard events it is paid in false rejects under an inherited threshold; without them the utility loss averages under 1~pp. The three audits separate the error, its origin and its cost. As $b^*$ does not depend on the model or its input, world models that read images share this optimum; whether they reach it in practice is open.

\textbf{Limitations.} The main results are simulated (real-robot experiments are in the supplement): one language model, one truth judge with a known channel, contact layouts only and no visual input; most settings have two feedback seeds, and regret rests on 26 to 49 layouts. Phantom mass depends on the enumerated reference: a 3.15 times denser grid lowers it by about 13 to 14\%, and the stability margin fixed in advance was not met. We do not train with a reference model or another $\beta$, as in DPO (Supplementary \S A.3), nor Full Channel with an estimated judge. Whether such feedback can also cause underestimation, and with it false allows, is untested. The threshold loss vanishes when thresholds are re-tuned on truth, so its weight depends on the cost of truth.

\section*{LLM Usage Statement}
Anthropic Claude was used to assist with grammar correction and improving the clarity of the paper. All suggested edits were reviewed by the authors, who take full responsibility for the final content.

{
    \small
    \interlinepenalty=10000
    \bibliographystyle{ieeenat_fullname}
    \bibliography{references}
}
\end{document}